# Cumulative distribution networks and the derivative-sum-product algorithm

Jim C. Huang   and   Brendan J. Frey
Probabilistic and Statistical Inference Group, University of Toronto
10 King's College Road, Toronto, ON
M5S 3G4, Canada

## Abstract

We introduce a new type of graphical model called a 'cumulative distribution network' (CDN), which expresses a joint cumulative distribution as a product of local functions. Each local function can be viewed as providing evidence about possible orderings, or rankings, of variables. Interestingly, we find that the conditional independence properties of CDNs are quite different from other graphical models. We also describe a message-passing algorithm that efficiently computes conditional cumulative distributions. Due to the unique independence properties of the CDN, these messages do not in general have a one-to-one correspondence with messages exchanged in standard algorithms, such as belief propagation. We demonstrate the application of CDNs for structured ranking learning using a previously-studied multi-player gaming dataset.

## 1 Introduction

Probabilistic graphical models are widely used for compactly representing joint probability density functions (PDFs)[1]. While such models have been successfully applied to a variety of problems, there are many tasks in which the joint probability density does not arise naturally and other representations may be more appropriate. In particular, the *cumulative distribution function* (CDF) is a probabilistic representation which arises frequently in a wide variety of applications such as ranking learning [9], survival analysis and data censuring [3]. In the setting of ranking learning, the goal is to model an ordinal variable $y$ given an input set of features $\mathbf{x}$ while accounting for noise in the ranking process. The conditional CDF here accounts for the ordinal nature of $y$ in addition to model uncertainty.

[1]We use PDF in reference to either the probability density function in the case of continuous random variables or the probability mass function in the case of discrete random variables

For problems of ranking learning that exhibit structure in the form of predicting multiple ordinal variables, a more flexible representation is required. Examples of this type of problem include predicting movie ratings for multiple movies or predicting multiplayer game outcomes with a team structure [4]. In such settings, we may wish to model not only stochastic ordering relationships between variable states, but also preferences between model variables and stochastic independence relationships between variables. This requires representing the joint CDF of many variables whilst explicitly accounting for both marginal and conditional independencies between variables.

Motivated by the above two problems, we present the cumulative distribution network (CDN), a novel graphical model which describes the joint CDF of a set of variables instead of the joint PDF. We show that CDNs provide a compact way to represent stochastic ordering relationships amongst variables and that the conditional independence properties of CDNs are quite different from previously studied graphical models describing PDFs (Bayesian networks [12], Markov random fields [6], factor graphs [7] and chain graphs [1]). In contrast to those, marginalization in CDNs involves tractable operations such as computing derivatives of local functions. We derive relevant theorems and lemmas for CDNs and describe a message-passing algorithm called the *derivative-sum-product algorithm* (DSP) for performing inference in such models. Finally we present results on an application to structured ranking learning in a multiplayer online game setting.

### 1.1 Example: Expressing conditional dependencies between variables

To illustrate a situation in which we have a set of conditional dependence relationships that cannot be represented by either a Markov random field or a Bayesian network, consider the following probability model with 4 binary variables $X_1, X_2, X_3, X_4$.

From the above probability model, we can establish

| $x_1$ | $x_2$ | $x_3$ | $x_4$ | $P(x_1,x_2,x_3,x_4)$ |
|---|---|---|---|---|
| 0 | 0 | 0 | 0 | 343/1800 |
| 0 | 0 | 0 | 1 | 392/1800 |
| 0 | 0 | 1 | 0 | 105/1800 |
| 0 | 0 | 1 | 1 | 168/1800 |
| 0 | 1 | 0 | 0 | 105/1800 |
| 0 | 1 | 0 | 1 | 120/1800 |
| 0 | 1 | 1 | 0 | 87/1800 |
| 0 | 1 | 1 | 1 | 120/1800 |
| 1 | 0 | 0 | 0 | 49/1800 |
| 1 | 0 | 0 | 1 | 56/1800 |
| 1 | 0 | 1 | 0 | 15/1800 |
| 1 | 0 | 1 | 1 | 24/1800 |
| 1 | 1 | 0 | 0 | 63/1800 |
| 1 | 1 | 0 | 1 | 72/1800 |
| 1 | 1 | 1 | 0 | 33/1800 |
| 1 | 1 | 1 | 1 | 48/1800 |

Table 1: Example probability model over 4 binary variables $X_1, X_2, X_3, X_4$.

that $X_1 \not\!\perp\!\!\!\perp X_3 | X_2$ (i.e.: $X_1$ is dependent of $X_3$ given $X_2$), as $P(x_1, x_3|x_2) \neq P(x_1|x_2)P(x_3|x_2)$:

|  | $P(x_1,x_3\|x_2)$ | | $P(x_1\|x_2)P(x_3\|x_2)$ | |
|---|---|---|---|---|
| $x_1, x_3$ | $x_2 = 0$ | $x_2 = 1$ | $x_2 = 0$ | $x_2 = 1$ |
| 0,0 | 245/384 | 75/216 | 245/384 | 80/216 |
| 0,1 | 91/384 | 69/216 | 91/384 | 64/216 |
| 1,0 | 35/384 | 45/216 | 35/384 | 40/216 |
| 1,1 | 13/384 | 27/216 | 13/384 | 32/216 |

One can also verify that $X_2 \not\!\perp\!\!\!\perp X_4|X_3$, $X_1 \not\!\perp\!\!\!\perp X_2$, $X_2 \not\!\perp\!\!\!\perp X_3$, $X_3 \not\!\perp\!\!\!\perp X_4, X_1 \perp\!\!\!\perp X_4, X_1 \perp\!\!\!\perp X_3, X_2 \perp\!\!\!\perp X_4$ from the above joint probability. In fact, we cannot represent the above set of conditional independence and dependence relationships using either a Markov random field, a Bayesian network or a factor graph, as $X_1 \not\!\perp\!\!\!\perp X_3|X_2$ and $X_1 \perp\!\!\!\perp X_3$ cannot be simultaneously satisfied under such frameworks (similarly for $X_2 \not\!\perp\!\!\!\perp X_4|X_3$ and $X_2 \perp\!\!\!\perp X_4$). Attempting to construct a directed model would lead to the introduction of directed cycles (Figure 1(a)) and would not be valid as a Bayesian network. However, the probability model in this example can be represented in compact form as a CDN (Figure 1(b)), which we will now proceed to define.

## 2 Cumulative distribution networks (CDNs)

Let $\mathbf{X} = \{X_1, \cdots, X_K\}$ denote a collection of $K$ real-valued ordinal random variables. Let $\mathbf{x} = (x_1, \cdots, x_K)$ denote a vector of assignments for these variables under a joint PDF given by $P(\mathbf{x}) =$ $P(x_1, \cdots, x_K) = Pr\{X_1 = x_1, \cdots, X_K = x_K\}$. Similarly, let $\mathbf{X} \setminus X_k$ denote the set $\mathbf{X}$ with $X_k$ deleted. Let $\mathbf{X}_s$ denote some subset of the set of variables $\mathbf{X}$. We will often resort to the differentiation or finite difference operator with respect to a set of variables $\mathbf{X}_s$: we will use the notation $\partial_{\mathbf{x}_s}\left[\cdot\right]$ to denote this.

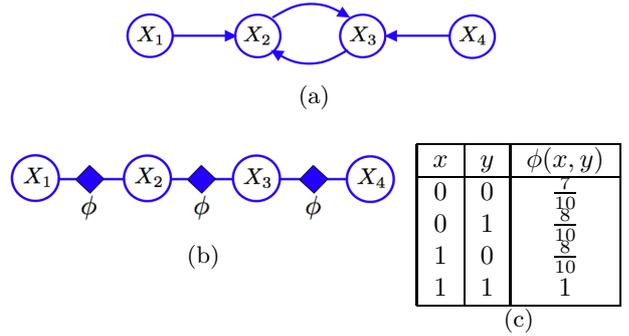

Figure 1: a) Attempting to construct a directed model that encodes all conditonal independence and dependence relationships in the model of Table 1 leads to directed cycles, which are improper; b) The corresponding cumulative distribution network (CDN) over the 4 variables $X_1, X_2, X_3, X_4$; c) Function $\phi(x, y)$ used to construct the CDN in Table 1.

Our analysis here assumes the existence of the joint PDF $P(\mathbf{x})$ over a set of variables. The joint CDF $F(\mathbf{x}) = F(x_1, \cdots, x_K) = Pr\{X_1 \leq x_1, \cdots, X_K \leq x_K\}$ is defined as a function over $\{x_1, \cdots, x_K\}$ such that $\partial_{\mathbf{x}}\left[F(\mathbf{x})\right] = P(\mathbf{x})$ and

$$\sup_{\mathbf{x}} F(\mathbf{x}) = 1 \qquad (1)$$

$$\inf_{x_k} F(x_k, \mathbf{x} \setminus x_k) = 0 \quad \forall k = 1, \cdots, K \qquad (2)$$

$$\partial_{\mathbf{x}_s}\left[F(\mathbf{x}_s, \mathbf{x} \setminus \mathbf{x}_s)\right] \geq 0 \quad \forall \mathbf{X}_s \subseteq \mathbf{X} \qquad (3)$$

We will now define the concept of a cumulative distribution network as a graphical model that efficiently expresses a CDF as a product of local functions, each of which captures a local cumulative distribution function.

**Definition.** *For $K$ random variables $X_1, \cdots, X_K$, the cumulative distribution network (CDN) is an undirected bipartite graphical model consisting of a set of variable nodes corresponding to variables in $\mathbf{X}$, cumulative function nodes defined over subsets of these variables, a set of undirected edges linking variables to functions and a specification for each cumulative function. More precisely, let $\mathcal{C}$ denote the set of cumulative functions so that for $c \in \mathcal{C}$, $\phi_c(\mathbf{x}_c)$ is a cumulative function defined over neighboring variables $\mathbf{X}_c$. Let $\mathrm{N}(X_i) = \{c \in \mathcal{C}|X_i \in \mathbf{X}_c\}$ denote the set of all neighboring functions for variable $X_i$ and let $\mathrm{N}(\phi_c) = \mathbf{X}_c$ be the set of variable nodes that $\phi_c$ is connected to. Simi-* larly, let $N(\mathbf{X}_s) = \bigcup_{X_i \in \mathbf{X}_s} N(X_i)$ denote the set of all neighboring functions for variable set $\mathbf{X}_s$. For a CDN with $K$ variable nodes, the joint CDF of $X_1, \cdots, X_K$ is given by

$$F(\mathbf{x}) = F(x_1, \cdots, x_K) = \prod_{c \in \mathcal{C}} \phi_c(\mathbf{x}_c). \qquad (4)$$

As an example CDN, consider a joint CDF over 3 variables $X, Y, Z$ which can be expressed as

$$F(x, y, z) = \phi_a(x, y) \phi_b(x, y, z) \phi_c(y, z) \phi_d(z). \qquad (5)$$

This can be represented using the undirected graph shown in Figure 2, where each function node corresponds to one of the functions $\phi_c(\mathbf{x}_c)$.

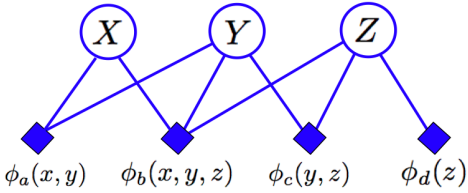

Figure 2: A cumulative distribution network (CDN) over 3 variables and 4 functions.

In order for the CDN to describe a joint CDF that corresponds to a valid PDF, we require that each of the cumulative functions satisfy $\phi_c(\mathbf{x}_c) \geq 0$. Furthermore, we will assume in the sequel that the derivatives/finite differences of each $\phi_c(\mathbf{x}_c)$ with respect to all of its argument variables can be properly computed. Finally, to satisfy the basic properties (1) to (3), the functions $\phi_c(\mathbf{x}_c)$ must satisfy the following conditions.

**Positive convergence** For all $c \in \mathcal{C}$, (1) holds if

$$\sup_{\mathbf{x}_c} \phi_c(\mathbf{x}_c) = 1. \qquad (6)$$

We note that the former condition (6) is a sufficient, but not a necessary condition for (1) to hold, as we could allow for functions to converge to different limits whilst retaining the property that $\sup_{\mathbf{x}} F(\mathbf{x}) = 1$.

**Negative convergence** For any $X_i \in \mathbf{X}$, (2) holds if and only if at least one of its neighboring functions $\phi_c(\mathbf{x}_c)$ goes to zero as $x_i$ goes to zero, i.e.

$$\forall X_i, \exists c \in N(X_i) \mid \inf_{x_i} \phi_c(\mathbf{x}_c) = 0. \qquad (7)$$

This can be proven as follows. For any $X_i \in \mathbf{X}$, $c \in N(X_i)$, if $\inf_{x_i} \phi_c(\mathbf{x}_c) = 0$, then $\inf_{x_i} F(\mathbf{x}) = 0$, as

$$\inf_{x_i} F(\mathbf{x}) = \inf_{x_i} \prod_{c \in N(X_i)} \phi_c(x_i, \mathbf{x}_c \setminus x_i) \prod_{c \notin N(X_i)} \phi_c(\mathbf{x}_c)$$

$$= \prod_{c \notin N(X_i)} \phi_c(\mathbf{x}_c) \prod_{c \in N(X_i)} \inf_{x_i} \phi_c(x_i, \mathbf{x}_c \setminus x_i)$$

$$= 0.$$

Conversely,

$$\inf_{x_i} F(\mathbf{x}) = \inf_{x_i} \prod_{c \in N(X_i)} \phi_c(x_i, \mathbf{x}_c \setminus x_i) \prod_{c \notin N(X_i)} \phi_c(\mathbf{x}_c) = 0$$

$$\Rightarrow \exists c \in N(X_i) \mid \inf_{x_i} \phi_c(\mathbf{x}_c) = 0.$$

**Monotonicity** For any $\mathbf{x}$ such that $F(\mathbf{x}) > 0$ and for any $X_i \in \mathbf{X}$, $\partial_{x_i} [F(\mathbf{x})] \geq 0$ holds if and only if

$$\sum_{c \in N(X_i)} \frac{\partial_{x_i} [\phi_c(\mathbf{x}_c)]}{\phi_c(\mathbf{x}_c)} \geq 0. \qquad (8)$$

To prove this, since $F(\mathbf{x}) = \prod_{c \in \mathcal{C}} \phi_c(\mathbf{x}_c)$, we have

$$\partial_{x_i} [F(\mathbf{x})] = \partial_{x_i} \Big[ \prod_{c \in N(X_i)} \phi_c(\mathbf{x}_c) \prod_{c \notin N(X_i)} \phi_c(\mathbf{x}_c) \Big]$$

$$= \prod_{c \in \mathcal{C}} \phi_c(\mathbf{x}_c) \sum_{c \in N(X_i)} \frac{1}{\phi_c(\mathbf{x}_c)} \partial_{x_i} [\phi_c(\mathbf{x}_c)]$$

$$= F(\mathbf{x}) \sum_{c \in N(X_i)} \frac{1}{\phi_c(\mathbf{x}_c)} \partial_{x_i} [\phi_c(\mathbf{x}_c)] \geq 0$$

$$\Rightarrow \sum_{c \in N(X_i)} \frac{\partial_{x_i} [\phi_c(\mathbf{x}_c)]}{\phi_c(\mathbf{x}_c)} \geq 0.$$

Conversely,

$$\sum_{c \in N(X_i)} \frac{1}{\phi_c(\mathbf{x}_c)} \partial_{x_i} [\phi_c(\mathbf{x}_c)] = \partial_{x_i} \Big[ \log \prod_{c \in N(X_i)} \phi_c(\mathbf{x}_c) \Big]$$

$$= \partial_{x_i} \Big[ \log \frac{F(\mathbf{x})}{\prod_{c \notin N(X_i)} \phi_c(\mathbf{x}_c)} \Big] = \partial_{x_i} \Big[ \log F(\mathbf{x}) \Big]$$

$$= \frac{1}{F(\mathbf{x})} \partial_{x_i} [F(\mathbf{x})] \geq 0 \Rightarrow \partial_{x_i} [F(\mathbf{x})] \geq 0.$$

The above result points to a sufficient condition for the CDN to correspond to a valid PDF, as demonstrated by the following lemma.

**Lemma** For all $\phi_c, X_i \in N(\phi_c)$, $\partial_{x_i} [F(\mathbf{x})] \geq 0$ if $\partial_{x_i} [\phi_c(\mathbf{x}_c)] \geq 0$. To prove this, since $\phi_c(\mathbf{x}_c) \geq 0 \, \forall c \in \mathcal{C}$, we see that $\partial_{x_i} [\phi_c(\mathbf{x}_c)] \geq 0$ immediately satisfies $\sum_{c \in N(X_i)} \frac{\partial_{x_i} [\phi_c(\mathbf{x}_c)]}{\phi_c(\mathbf{x}_c)} \geq 0$.

More generally, the above also holds for derivatives with respect to subsets of variables $\mathbf{x}_s$. That is, for all $c \in \mathcal{C}$, $\mathbf{X}_c = N(\phi_c)$ and all possible subsets of variables $\mathbf{X}_s \subseteq \mathbf{X}_c$ such that $F(\mathbf{x}_s, \mathbf{x} \setminus \mathbf{x}_s)$ is strictly positive, $\partial_{\mathbf{x}} [F(\mathbf{x})] \geq 0$ holds if $\partial_{\mathbf{x}_s} [\phi_c(\mathbf{x}_c)] \geq 0$. This can be proven by noting that computing higher-order derivatives of $F(\mathbf{x})$ yields sums over products of derivatives, so if $\phi_c(\mathbf{x}_c)$ and all higher-order derivatives with respect to each of their arguments are non-negative, then the resulting joint CDF must be monotonically non-decreasing and so (3) is satisfied.

## 3 Marginal and conditional independence properties of CDNs

In this section, we will highlight the marginal and conditional independence properties of CDNs, the latter of which significantly differ from those of graphical density models such as Bayesian networks, Markov random fields and factor graphs.

**Theorem (Marginal Independence).** *Two disjoint sets of variables $\mathbf{X}_s$ and $\mathbf{X}_t$ are marginally independent if there are no cumulative functions $c \in \mathcal{C}$ connecting variable nodes in $\mathbf{X}_s$ and $\mathbf{X}_t$.*

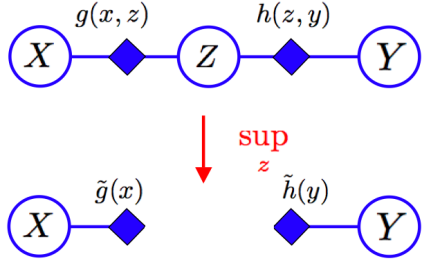

Figure 3: Marginal independence property of CDNs: unless two variables $X$ and $Y$ are connected to a common function node, they are marginally independent.

*Proof.* Intuitively, $\mathbf{X}_s$ and $\mathbf{X}_t$ are marginally independent if they have no neighboring functions in common. We can thus write
$$F(\mathbf{x}) = \prod_{c \in N(\mathbf{X}_s)} \phi_c(\mathbf{x}_c) \prod_{c \in N(\mathbf{X}_t)} \phi_c(\mathbf{x}_c) \prod_{c \notin N(\mathbf{X}_s) \bigcup N(\mathbf{X}_t)} \phi_c(\mathbf{x}_c).$$
(9)
Marginalizing over other variables $\mathbf{X} \setminus \{\mathbf{X}_s, \mathbf{X}_t\}$, we obtain
$$F(\mathbf{x}_s, \mathbf{x}_t) = \sup_{\mathbf{x} \setminus \{\mathbf{x}_s, \mathbf{x}_t\}} F(\mathbf{x})$$
$$= \left( \sup_{\mathbf{x} \setminus \mathbf{x}_s} \prod_{c \in N(\mathbf{X}_s)} \phi_c(\mathbf{x}_c) \right) \left( \sup_{\mathbf{x} \setminus \mathbf{x}_t} \prod_{c \in N(\mathbf{X}_t)} \phi_c(\mathbf{x}_c) \right)$$
$$\cdot \left( \sup_{\mathbf{x} \setminus \{\mathbf{x}_s, \mathbf{x}_t\}} \prod_{c \notin N(\mathbf{X}_s) \bigcup N(\mathbf{X}_t)} \phi_c(\mathbf{x}_c) \right)$$
$$= \left( \sup_{\mathbf{x} \setminus \mathbf{x}_s} \prod_{c \in N(\mathbf{X}_s)} \phi_c(\mathbf{x}_c) \right) \left( \sup_{\mathbf{x} \setminus \mathbf{x}_t} \prod_{c \in N(\mathbf{X}_t)} \phi_c(\mathbf{x}_c) \right)$$
$$= g(\mathbf{x}_s) h(\mathbf{x}_t),$$
where $\sup_{\mathbf{x}_s} g(\mathbf{x}_s) = \sup_{\mathbf{x}_t} h(\mathbf{x}_t) = 1$ follows from (6), so $F(\mathbf{x}_s) = g(\mathbf{x}_s)$ and $F(\mathbf{x}_t) = h(\mathbf{x}_t)$. Thus, we have $F(\mathbf{x}_s, \mathbf{x}_t) = F(\mathbf{x}_s)F(\mathbf{x}_t)$ and so $\mathbf{X}_s \perp\!\!\!\perp \mathbf{X}_t$. □

An example of the marginal independence property for a 3-variable CDN is shown in Figure 3. Another interesting property of CDNs is that conditioning on variables in the network corresponds to computing derivatives or finite differences of the joint CDF, as we will now show.

**Theorem (Conditioning).** *Conditioned on a set of variables $\mathbf{X}_s = \mathbf{x}_s$, we have*
$$F(\mathbf{x}_t | \mathbf{x}_s) = \frac{\partial_{\mathbf{x}_s}\big[F(\mathbf{x}_t, \mathbf{x}_s)\big]}{\sup_{\mathbf{x}_t} \partial_{\mathbf{x}_s}\big[F(\mathbf{x}_t, \mathbf{x}_s)\big]} \propto \partial_{\mathbf{x}_s}\big[F(\mathbf{x}_t, \mathbf{x}_s)\big].$$
(10)

*Proof.* Note that $\partial_{\mathbf{x}_s}\big[F(\mathbf{x}_t, \mathbf{x}_s)\big] = F(\mathbf{x}_t | \mathbf{x}_s) P(\mathbf{x}_s) = F(\mathbf{x}_t | \mathbf{x}_s) \sup_{\mathbf{x}_t} \partial_{\mathbf{x}_s}\big[F(\mathbf{x}_t, \mathbf{x}_s)\big]$ and so
$$F(\mathbf{x}_t | \mathbf{x}_s) = \frac{\partial_{\mathbf{x}_s}\big[F(\mathbf{x}_t, \mathbf{x}_s)\big]}{\sup_{\mathbf{x}_t} \partial_{\mathbf{x}_s}\big[F(\mathbf{x}_t, \mathbf{x}_s)\big]}.$$
□

Equation (10) establishes the key operation of conditioning in CDNs: there is one-to-one correspondence between conditioning on a set of variables and computing the derivative/finite difference of the joint CDF with respect to these variables. This is a property we will exploit shortly when we are confronted with the problem of performing inference in a CDN.

**Theorem (Conditional Independence).** *Let $\mathbf{X}_s, \mathbf{X}_t$ and $\mathbf{X}_u$ be three disjoint sets of variables with no cumulative functions $c \in \mathcal{C}$ connecting any two variables in $\mathbf{X}_s, \mathbf{X}_t$. Then $\mathbf{X}_s \perp\!\!\!\perp \mathbf{X}_t | \mathbf{X}_u$ if every path between any two variables in $\mathbf{X}_s$ and $\mathbf{X}_t$ contains no variables in $\mathbf{X}_s, \mathbf{X}_t, \mathbf{X}_u$.*

*Proof.* We can marginalize over other variables $\mathbf{X} \setminus \{\mathbf{X}_s, \mathbf{X}_t, \mathbf{X}_u\}$. If this results in a CDN in which there is no path between any two variables in $\mathbf{X}_s$ and $\mathbf{X}_t$ such that all variable nodes in that path are in $\mathbf{X}_u$, then the sets $\{\mathbf{X}_s, \mathbf{X}_a\}, \{\mathbf{X}_t, \mathbf{X}_b\}$ consist of two disjoint subgraphs, with no cumulative functions $\phi_c$ connecting the two subgraphs and $\mathbf{X}_a, \mathbf{X}_b$ being two disjoint partitions of $\mathbf{X}_u$. Thus we can write
$$F(\mathbf{x}_s, \mathbf{x}_t, \mathbf{x}_u) = \prod_{c \in N(\mathbf{X}_a) \bigcup N(\mathbf{X}_s)} \phi_c(\mathbf{x}_c) \prod_{c \in N(\mathbf{X}_b) \bigcup N(\mathbf{X}_t)} \phi_c(\mathbf{x}_c)$$
$$\Rightarrow F(\mathbf{x}_s, \mathbf{x}_t | \mathbf{x}_u) \propto \partial_{\mathbf{x}_u}\big[F(\mathbf{x}_s, \mathbf{x}_t, \mathbf{x}_u)\big]$$
$$= \tilde{g}(\mathbf{x}_s, \mathbf{x}_a) \tilde{h}(\mathbf{x}_t, \mathbf{x}_b)$$
and so $\mathbf{X}_s \perp\!\!\!\perp \mathbf{X}_t | \mathbf{X}_u$. □

In summary, we have $\mathbf{X}_s \perp\!\!\!\perp \mathbf{X}_t | \mathbf{X}_u$ if all paths linking $\mathbf{X}_s$ to $\mathbf{X}_t$ are blocked by unobserved variables not in $\mathbf{X}_s, \mathbf{X}_t, \mathbf{X}_u$. An example of conditional dependence and independence is given in Figure 4.

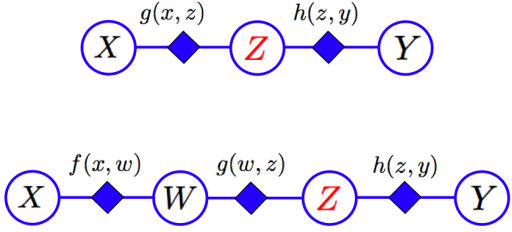

Figure 4: Conditional independence property of CDNs. Two variables $X$ and $Y$ become conditionally dependent given that the variable $Z$ is observed and no unobserved variables block the path from $X$ to $Y$ (*top*). When an unobserved variable $W$ blocks the path, $X$ and $Y$ are conditionally independent given $Z$ (*bottom*).

## 4 The derivative-sum-product algorithm for inference in cumulative distribution networks

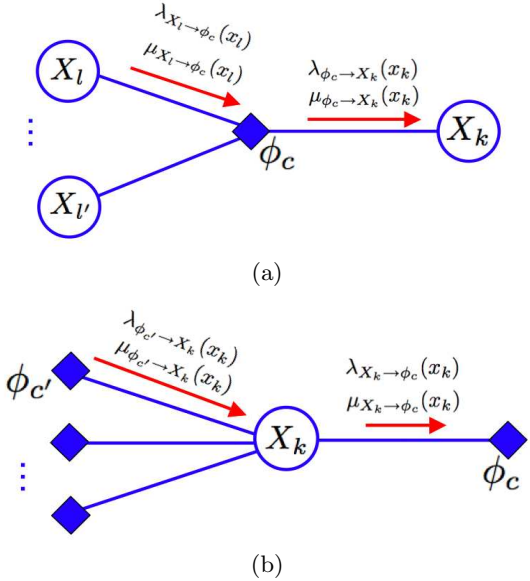

Figure 5: Messages in the derivative-sum-product algorithm: a) Computation of the messages $\mu_{\phi_c \to X_k}(x_k), \lambda_{\phi_c \to X_k}(x_k)$ from a function node $\phi_c$ to a variable node $X_k$; b) Computation of the messages $\mu_{X_k \to \phi_c}(x_k), \lambda_{X_k \to \phi_c}(x_k)$ from a variable node $X_k$ to a function node $\phi_c$

In this section we derive a *derivative-sum-product* algorithm for computing conditional CDFs efficiently in CDNs. We assume that the CDN has a tree structure to allow for exact inference, with the root variable node of the tree denoted by $X_r$. We also assume that the functions $\phi_c(\mathbf{x}_c)$ obey all of the necessary and sufficient conditions presented above: in particular we assume that all the functions $\phi_c$ in the network are differentiable with respect to all of their arguments. In the case in which variables are discrete, we can interchange differentiation with the simpler operation of finite differences. Without loss of generality, we can marginalize over all unobserved variables in the CDN by deleting them from the network and modifying their neighboring cumulative functions appropriately: thus, we assume that the network consists of fully observed variables. In the case we are differentiating (or computing finite differences) with respect to a set of variables which are observed with values $\mathbf{x}_s$, we implicitly assume that the resulting derivative is evaluated at the observed scalar values $\mathbf{x}_s$.

Before proceeding, it is worth asking whether there is a connection between message-passing in a CDN and message-passing in a density model such as a Bayesian network, Markov random field or a factor graph. As noted in Section 1.1, CDNs in general will not have a one-to-one corresponding representation in one of these graphical forms due to the unique conditional independence properties of the CDN. Thus in general, messages passed by the DSP algorithm will therefore be distinct from those exchanged under standard message-passing algorithms such as Pearl's belief propagation [12] or the sum-product algorithm [7] and will not generally correspond to computing the same sufficient statistics.

Our derivation of the DSP algorithm is analogous to that of Pearl's belief propagation algorithm [12]. The intuition here is similar in that we can distribute the differentiation operation to local functions such that each one computes the derivatives with respect to local variables and passes the result to its neighbors in the form of messages $\mu_{\phi_c \to X_k}(x_k)$ from function nodes to variables nodes and $\mu_{X_k \to \phi_c}(x_k)$ from variable nodes to function nodes. Since we assume that the derivatives/finite differences of the functions $\phi_c(\mathbf{x}_c)$ can be computed, we can apply the product rule of differential calculus to expand the messages in the DSP algorithm in terms of derivatives of the functions $\phi_c(\mathbf{x}_c)$ and derivatives of the messages. If we denote $\partial_{x_k}\left[\mu_{\phi_c \to X_k}(x_k)\right] = \lambda_{\phi_c \to X_k}(x_k)$ and $\partial_{x_k}\left[\mu_{X_k \to \phi_c}(x_k)\right] = \lambda_{X_k \to \phi_c}(x_k)$, we can write DSP using two sets of messages $\mu$ and $\lambda$, as shown in Figure 6.

The joint PDF is then given at the root node $X_r$ by $P(\mathbf{x}) = \lambda_{X_r \to \emptyset}(x_r)$. We need only to pre-compute the derivatives of the $\phi_c$'s with respect to all their arguments and combine these locally. Note that to compute the joint PDF $P(\mathbf{x})$ from $F(\mathbf{x})$ by brute force for $K$ variables over $N$ functions would require a summa-

- For each leaf variable node $X_l$ and for all functions $c \in \mathrm{N}(X_l)$, propagate $\mu_{X_l \to \phi_c}(x_l) = \lambda_{X_l \to \phi_c}(x_l) = 1$. For each leaf function node $\phi_l(x_l)$, send the messages $\mu_{\phi_l \to X_l}(x_l) = \phi_l(x_l), \lambda_{\phi_l \to X_l}(x_l) = \partial_{x_l}\left[\phi_l(x_l)\right]$.

- For each non-leaf variable $X_k$ and neighboring functions $c \in \mathrm{N}(X_k)$,

$$\mu_{X_k \to \phi_c}(x_k) = \prod_{c' \in \mathrm{N}(X_k) \setminus c} \mu_{\phi_{c'} \to X_k}(x_k) \tag{11}$$

$$\lambda_{X_k \to \phi_c}(x_k) = \partial_{x_k}\left[\mu_{X_k \to \phi_c}(x_k)\right] = \mu_{X_k \to \phi_c}(x_k) \sum_{c' \in \mathrm{N}(X_k) \setminus c} \frac{\lambda_{\phi_{c'} \to X_k}(x_k)}{\mu_{\phi_{c'} \to X_k}(x_k)} \tag{12}$$

- For each non-leaf function $c$ and neighboring variables $X_k \in \mathrm{N}(\phi_c)$,

$$\mu_{\phi_c \to X_k}(x_k) = \sum_{\substack{s,t \mid \mathbf{X}_s \bigcup \mathbf{X}_t = \mathbf{X}_c \setminus X_k \\ \mathbf{X}_s \bigcap \mathbf{X}_t = \emptyset}} \partial_{\mathbf{x}_s}\left[\phi_c(\mathbf{x}_c)\right] \prod_{j \mid X_j \in \mathbf{X}_s} \mu_{X_j \to \phi_c}(x_j) \prod_{j \mid X_j \in \mathbf{X}_t} \lambda_{X_j \to \phi_c}(x_j) \tag{13}$$

$$\lambda_{\phi_c \to X_k}(x_k) = \partial_{x_k}\left[\mu_{\phi_c \to X_k}(x_k)\right]$$

$$= \sum_{\substack{s,t \mid \mathbf{X}_s \bigcup \mathbf{X}_t = \mathbf{X}_c \setminus X_k \\ \mathbf{X}_s \bigcap \mathbf{X}_t = \emptyset}} \partial_{\mathbf{x}_s, x_k}\left[\phi_c(\mathbf{x}_c)\right] \prod_{j \mid X_j \in \mathbf{X}_s} \mu_{X_j \to \phi_c}(x_j) \prod_{j \mid X_j \in \mathbf{X}_t} \lambda_{X_j \to \phi_c}(x_j) \tag{14}$$

- Repeat the $2^{nd}$ and $3^{rd}$ steps above from root to leaf nodes and fuse messages at each variable node to get the conditionals.

Figure 6: The derivative-sum-product message-passing algorithm for computing conditionals in a cumulative distribution network.

tion with a number of product terms of $O(N^K)$. However with derivative-sum-product, we have reduced this to $O(\max\{\max_v(deg(v) - 1), \max_\phi(2^{deg(\phi)-1})\})$, where $v$ represents any variable node in the CDN and $\phi$ represents any function node.

## 5 Application: Structured ranking learning using CDNs

In structured ranking learning, we are interested in performing inference on multiple ordinal variables subject to a set of ordinal and stochastic independence/dependence relationships. While it is possible to formulate the problem under the framework of graphical density models, it is quite easy to specify both ordinal and *statistical* independence/dependence relationships between ordinal variables using the CDN framework while allowing for both a flexible model and tractable inference. To demonstrate the application of CDNs to structured ranking learning, we focus on the problem of rank prediction in multiplayer online gaming.

We downloaded the Halo 2 Beta Dataset[2] consisting of player performance scores for 4 game types ("LargeTeam","SmallTeam","HeadToHead' and

[2]Credits for the use of the Halo 2 Beta Dataset are given to Microsoft Research Ltd. and Bungie.

"FreeForAll") over a total of 6465 players and thousands of games. For each game, an outcome is defined by the performances of teams once the game is over, where team performances are determined by adding players performances together. Recently, [4] presented the TrueSkill algorithm for skill rating, whereby each player is assigned a distribution over skill levels which are inferred from individual player performances over multiple games using expectation propagation [10].

Our model is designed according to two principles: firstly, that the relationship between player scores and game outcomes is stochastic, and secondly, that game outcomes are determined by teams competing with one another on the basis of team performances. To address the first point, we will require a set of CDN functions which link player scores to team performance. Interpreting team performances as outputs, we will make use of the cumulative model [9], which relates a function $f(\mathbf{x})$ of the input variables $\mathbf{x}$ to the output ordinal variable $y \in \{r_1, \cdots, r_K\}$ so that $Pr\{y = r_i\} = Pr\{\theta(r_{i-1}) < f(\mathbf{x}) + \epsilon \leq \theta(r_i)\} = F_\epsilon(\theta(r_i) - f(\mathbf{x})) - F_\epsilon(\theta(r_{i-1}) - f(\mathbf{x}))$, where $\epsilon$ is additive noise and we define $\theta(r_0) = -\infty, \theta(r_K) = \infty$. Here, we will choose $\mathbf{x}$ as the set of player scores for a given team and so the team performance is given by $f(\mathbf{x}) = \mathbf{1}^T\mathbf{x}$. So if for any given game, there are $N$ teams, then each team is assigned a function $g_n$ such

that

$$g_n(\mathbf{x}_n, r_n) = \int_{-\infty}^{\mathbf{x}_n} F\big(\theta(r_n); \mathbf{1}^T\mathbf{u}, \beta^2\big) P(\mathbf{u})\, d\mathbf{u} \quad (15)$$

where $F\big(\theta(r_n); \mathbf{1}^T\mathbf{u}, \beta^2\big)$ is a cumulative model relating player scores to team performance, $\theta(r_n)$ are "cutpoints" which define contiguous intervals in which the team performance is assigned as $y_n = r_i$ and $P(\mathbf{u})$ is a density over the vector of player scores $\mathbf{u}$. We will model these functions as

$$F\big(\theta(r_n); \mathbf{1}^T\mathbf{u}, \beta^2\big) = \Phi\big(\theta(r_n); \mathbf{1}^T\mathbf{u}, \beta^2\big), \quad (16)$$
$$P(\mathbf{u}) = \mathcal{N}(\mathbf{u}; \mu\mathbf{1}, \sigma^2\mathbf{I}), \quad (17)$$

where $\Phi(\cdot; m, s^2)$ is the cumulative distribution function for a univariate Gaussian with mean $m$ and variance $s^2$, and $\mathcal{N}(\cdot; \boldsymbol{m}, \mathbf{S})$ is a multivariate Gaussian density.

To address the fact that teams compete for higher rank, we model ordinal relationships between team performance so that for two teams with performances $R_X, R_Y$, $R_X \preceq R_Y$ if $F_{R_X}(\alpha) \geq F_{R_Y}(\alpha)\,\forall\alpha$, where $F_{R_X}(\cdot), F_{R_Y}(\cdot)$ are the marginal CDFs of $R_X, R_Y$. Given $N$ ranked teams, we can thus define $N-1$ functions $h_{n,n+1}$ so that

$$h_{n,n+1}(r_n, r_{n+1}) = \Phi\left(\begin{bmatrix} r_n \\ r_{n+1} \end{bmatrix}; \begin{bmatrix} \tilde{r}_n \\ \tilde{r}_{n+1} \end{bmatrix}, \boldsymbol{\Sigma}\right)$$

where

$$\boldsymbol{\Sigma} = \begin{bmatrix} \beta^2 & \rho\beta^2 \\ \rho\beta^2 & \beta^2 \end{bmatrix}$$

and $\tilde{r}_n \leq \tilde{r}_{n+1}$ so as to enforce $R_n \preceq R_{n+1}$ in the overall model. The CDN corresponding to our model is shown in Figure 7(a): one can verify for any given game that indeed the relationship $R_1 \preceq R_2 \preceq \cdots \preceq R_N$ is enforced by marginalizing over player scores. Our model includes a function $s_k(x_k)$ for each player $k$ which represents that player's skill. The goal is then to learn the player skills: we used an online learning method for updating the functions $s_k(x_k)$, whereby for each game we perform message-passing followed by the update $s_k(x_k) \leftarrow s_k(x_k)\mu_{g_n \to X_k}(x_k)$ for each player. Before each game, we predict the outcome using the player skills learned thus far via $x_k^* = \arg\max_{x_k} \partial_{x_k}\big[s_k(x_k)\big] = \arg\max_{x_k} \lambda_{X_k \to s_k}(x_k)$.

Since all of the CDN functions are themselves multivariate Gaussian CDFs, the derivatives $\partial_{\mathbf{x}_s}\big[\phi_c(\mathbf{x}_c)\big]$ in the corresponding message-passing algorithm can be easily evaluated w.r.t. variables $\mathbf{X}_s$ as

$$\partial_{\mathbf{x}_s}\big[\Phi(\mathbf{x}; \boldsymbol{\mu}, \boldsymbol{\Sigma})\big] = \mathcal{N}\big(\mathbf{x}_s; \boldsymbol{\mu}_s, \boldsymbol{\Sigma}_s\big)\Phi\big(\mathbf{x}_t; \tilde{\boldsymbol{\mu}}_t, \tilde{\boldsymbol{\Sigma}}_t\big)$$

where

$$\mathbf{x} = \begin{bmatrix} \mathbf{x}_s \\ \mathbf{x}_t \end{bmatrix}, \quad \boldsymbol{\mu} = \begin{bmatrix} \boldsymbol{\mu}_s \\ \boldsymbol{\mu}_t \end{bmatrix}, \quad \boldsymbol{\Sigma} = \begin{bmatrix} \boldsymbol{\Sigma}_s & \boldsymbol{\Sigma}_{s,t} \\ \boldsymbol{\Sigma}_{s,t}^T & \boldsymbol{\Sigma}_t \end{bmatrix},$$

$$\tilde{\boldsymbol{\mu}}_t = \boldsymbol{\mu}_t + \boldsymbol{\Sigma}_{s,t}^T \boldsymbol{\Sigma}_s^{-1}(\mathbf{x}_s - \boldsymbol{\mu}_s),$$
$$\tilde{\boldsymbol{\Sigma}}_t = \boldsymbol{\Sigma}_t - \boldsymbol{\Sigma}_{s,t}^T \boldsymbol{\Sigma}_s^{-1} \boldsymbol{\Sigma}_{s,t}.$$

To learn the model, we set the cutpoints $\theta(r_n)$ in the above model using ordinal regression of team performances on team scores, with $\mu$ set to the minimum player score. A plot showing the total prediction error rate of our method is shown in Figure 7(b), along with the error rates reported by [4] for TrueSkill and ELO [2], which is a statistical rating system used in chess.

## 6 Discussion and future work

We proposed a novel class of probabilistic graphical models, the cumulative distribution network (CDN). We demonstrated that the CDN has distinct statistical independence properties from Bayesian networks, Markov random fields and factors graphs. This also differs from convolutional factor graphs [8], which replace products of functions with convolutions whilst retaining the same conditional independence and dependence structure, albeit in the Fourier domain.

The task of computing marginals in CDNs is achievable in constant time. We proposed the derivative-sum-product (DSP) algorithm for computing these conditional distributions and demonstrated it to be a distinct from standard message-passing algorithms such as Pearl's belief propagation or the sum-product algorithm. We focused here on the application of CDNs to the problem of ranking learning in a multiplayer gaming setting. A list of other potential applications (which is by no means exhaustive) would include ranking webpages, collaborative filtering, inference of censured data [3] and generalized hypothesis testing in which one could both model statistical dependencies between multiple hypotheses.

While we have presented the DSP algorithm for computing conditionals given a set of cumulative functions, we have not addressed here the issue of how to learn these cumulative functions from data. An obvious method would be to run derivative-sum-product to obtain the joint PDF and then maximize this with respect to model parameters, but this approach has yet to be worked out in detail. Another issue we have not addressed is how to perform inference in graphs with cycles: an interesting future direction would be to investigate exact or approximate methods for doing so and connections to methods in the literature [5, 11] for doing this in traditional graphical models.

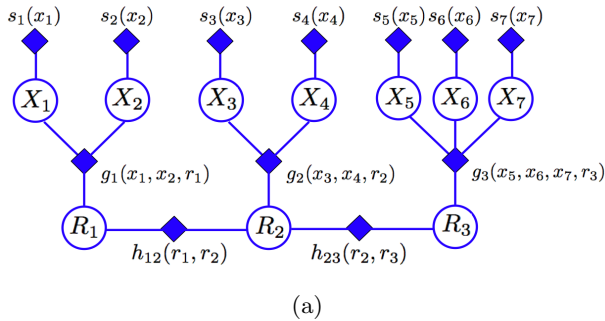 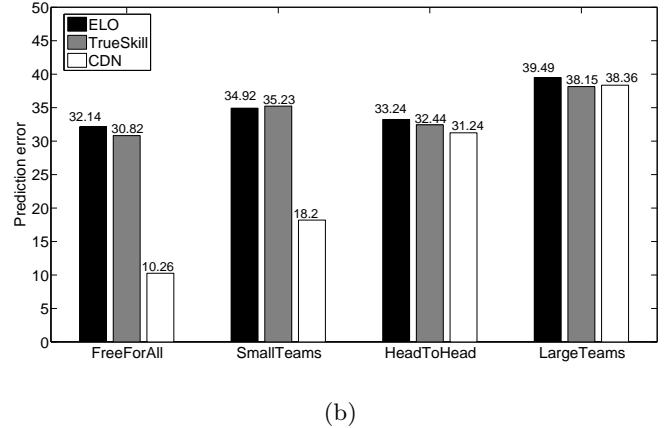

(a)          (b)

Figure 7: a) An example CDN corresponding to a multiplayer game with 3 teams of 2, 2 and 3 players with stochastic team rankings $R_1 \preceq R_2 \preceq R_3$. The first layer of functions $s_k(x_k)$ correspond to distributions over player performances (or *skills*). The functions $g_n, h_n$ relate team ranks to to one another as well as player performances; b) Prediction error (computed as the fraction of team predicted incorrectly before each game) achieved by our method, compared to ELO [2] and TrueSkill error [4].

## 7 Acknowledgements

We would like to thank anonymous reviewers, Radford Neal and Frank Kschischang for helpful comments. J.C.H. was supported by a National Science and Engineering Research Council postgraduate scholarship. This research was supported by an NSERC Discovery Grant.

## References

[1] Buntine, W. (1995) Chain graphs for learning. *In Proceedings of Conference on Uncertainty in Artificial Intelligence 11 (UAI 1995)*, 46-54.

[2] Elo, A.E. (1978) The rating of chess players: Past and present. *Arco Publishing, New York*.

[3] Gelman, A., Carlin, J.B., Stern, H.S. and Rubin, D.B. (2004) Bayesian data analysis, Second edition. *Texts in Statistical Science, Chapman & Hall*.

[4] Herbrich, R., Minka, T.P. and Graepel, T. (2007) TrueSkill$^{TM}$: A Bayesian skill rating system. *In Proceedings of Neural Information Processing Systems 2007 (NIPS 2007)*, 569-576.

[5] Jordan, M.I., Ghahramani, Z., Jaakkola, T.S. and Saul, L.K. (1999) An introduction to variational methods for graphical models. *Machine Learning* **37(2)**, 183-223.

[6] Kinderman, R. and Snell, J. L. (1980) Markov random fields and their applications. *American Mathematical Society*.

[7] Kschischang, F.R., Frey, B.J. and Loeliger, H.-A. (2001) Factor graphs and the sum-product algorithm. *IEEE Trans. Inform. Theory* **47(2)**, 498-519.

[8] Mao, Y., Kschischang, F.R. and Frey, B.J. (2004) Convolutional factor graphs as probabilistic models. *In Proceedings of Conference on Uncertainty in Artificial Intelligence 20 (UAI 2004)*, 374-381.

[9] McCullagh, P. (1980) Regression models for ordinal data. *J. Royal Statistical Society, Series B (Methodological)* **42(2)**, 109-142.

[10] Minka, T.P. (2001) Expectation propagation for approximate Bayesian inference. *In Proceedings of Conference on Uncertainty in Artificial Intelligence 17 (UAI 2001)*, 362-369.

[11] Neal, R. M. (1993) Probabilistic Inference Using Markov Chain Monte Carlo Methods, *Technical Report CRG-TR-93-1, Dept. of Computer Science, University of Toronto*, 144 pages.

[12] Pearl, J. (1988) Probabilistic reasoning in intelligent systems. *Morgan Kaufmann*.